\documentclass[
]{ceurart}

\usepackage{listings}
\usepackage{graphicx}
\usepackage{subcaption}
\usepackage{enumerate}
\begin{document}

\copyrightyear{2022}
\copyrightclause{Copyright for this paper by its authors.
 Use permitted under Creative Commons License Attribution 4.0
 International (CC BY 4.0).}

\conference{}

\title{Painting the black box white: experimental findings from applying XAI to an ECG reading setting}



\author[1,2]{Federico Cabitza}[%
email=federico.cabitza@unimib.it
]
\cormark[1]
\address[1]{Department of Computer Science, Systems and Communication, University of Milano-Bicocca, Milan, Italy}
\address[2]{IRCCS Istituto Ortopedico Galeazzi, Milan, Italy}

\author[3]{Matteo Cameli}
\address[3]{Department of Medicine, Surgery and Neuroscience, University of Siena, Siena, Italy}

\author[1]{Andrea Campagner}

\author[1]{Chiara Natali}

\author[4]{Luca Ronzio}
\address[4]{Department of Medicine and Surgery, University of Milano-Bicocca, Milan, Italy}

\cortext[1]{Corresponding author.}

\begin{abstract}
The shift from symbolic AI systems to black-box, sub-symbolic, and statistical ones has motivated a rapid increase in the interest toward explainable AI (XAI), i.e. approaches to make black-box AI systems explainable to human decision makers with the aim of making these systems more acceptable and more usable tools and supports. However, we make the point that, rather than always making black boxes transparent, these approaches are at risk of \emph{painting the black boxes white}, thus failing to provide a level of transparency that would increase the system's usability and comprehensibility; or, even, at risk of generating new errors, in what we termed the \emph{white-box paradox}.
To address these usability-related issues, in this work we focus on the cognitive dimension of users' perception of explanations and XAI systems. To this aim, we designed and conducted a questionnaire-based experiment by which we involved 44 cardiology residents and specialists in an AI-supported ECG reading task. In doing so, we investigated different research questions concerning the relationship between users' characteristics (e.g. expertise) and their perception of AI and XAI systems, including their trust, the perceived explanations' quality and their tendency to defer the decision process to automation (i.e. technology dominance), as well as the mutual relationships among these different dimensions. Our findings provide a contribution to the evaluation of AI-based support systems from a Human-AI interaction-oriented perspective and lay the ground for further investigation of XAI and its effects on decision making and user experience.

\end{abstract}

\maketitle

\section{Introduction}
We are witnessing a continuous and indeed accelerating move from decision support systems that are based on explicit rules conceived by domain experts (so called expert systems or knowledge-based systems) to systems whose behaviors can be traced back to an innumerable amount of rules that have been automatically learnt on the basis of correlative and statistical analyses of large quantities of data: this is the shift from symbolic AI systems to sub-symbolic ones, which has made the black-box nature of these latter systems an object of a lively and widespread debate in both technological and philosophical contexts~\cite{calegari2020integration}. The main assumption motivating this debate is that making subsymbolic systems explainable to human decision makers makes them better and more acceptable tools and supports. 

This assumption is widely accepted~\cite{cina2022we,gerlings2020reviewing,goebel2018explainable}, although there are a few scattered voices against it (see e.g.~\cite{de2022perils,janssen2022will,lipton2018mythos,schemmer2022influence}): for instance, explanations were found to increase complacency towards the machine advice~\cite{poursabzi2021manipulating}, increase automation bias~\cite{zhang2020effect,bansal2021does} as well as groundlessly increase confidence in one's own decision~\cite{eiband2019impact,ghassemi2021false}. Understanding or participating to this debate, which characterizes the scientific community that recognizes itself in the expression ``explainable AI'' and in the acronym ``XAI'', is difficult for the seemingly disarming heterogeneity of definitions of explanation, and the variety of characteristics that are associated with ``good explanations'', or of the systems that generate them~\cite{cabitza2022quod}. 

In what follows, we adopt the simplifying approach recently proposed in~\cite{cabitza2022quod}, where explanation is defined as the \emph{meta output} (that is an output that describes, enriches or complements, another main output) of an XAI systems. In this perspective, good explanations are those that make the XAI system more usable, and therefore a useful support. The reference to usability suggests that we can assess explanations (and explainability) on different levels, by addressing complementary questions, such as: do explanations make the socio-technical, decision-making setting more effective, in that they help decision makers commit fewer errors? Do they make it more efficient, by making decisions easier and faster, or just requiring fewer resources? And lastly, but not least, do they make users more satisfied with the advice received, possibly because they have understood it more, and this made them more confident about their final say? 

While some studies~\cite{shin2021effects} have already considered the psychometric dimension of user satisfaction (see, e.g., the concept of \emph{causability}~\cite{holzinger2019causability}, related to the role of explanations in making advice more understandable from a causal point of view), here we would like to focus on effectiveness (i.e., accuracy) and other cognitive dimensions (than understandability), both in regard to the support (e.g., trust and utility) and the explanations received. In fact, explanations can be either clear or ambiguous (cf. comprehensibility); either tautological and placebic~\cite{langer1978mindlessness} or instructive (cf. informativeness); either pertinent or off-topic (cf. pertinence); and, as obvious as it may seem, either correct or wrong, as any AI output can be: therefore, otherwise good explanations (that is persuasive, reassuring, comprehensible, etc.) could even mislead their target users: this is the so called \emph{white-box paradox}, which we have already begun investigating in previous empirical studies~\cite{cabitza2021need,cabitza2022color}. Thus, investigating if and how much users find explanations ``good'' (and in the next section we will make this term operationally clear) can be related to focusing on the possible determinants of machine influence (i.e., also called \emph{dominance}), automation bias and other negative effects related to the output of decision support systems on decision performance and practices. 

\section{Methods}
\label{sec:methods}

To investigate how human decision makers perceive explanations, we designed and conducted a questionnaire-based experiment in which we involved 44 cardiology of varying expertise and competence (namely, 25 residents and 19 specialists) from the Medicine School of the University Hospital of Siena (Italy), in an AI-supported ECG reading task, not connected to their daily care. The readers were invited to classify and annotate 20 ECG cases, previously selected by a cardiologist from a random set of cases extracted from the ECG Wave-Maven repository\footnote{\url{https://ecg.bidmc.harvard.edu/maven/mavenmain.asp}} on the basis of their complexity (recorded in the above repository), so as to have a balanced dataset in terms of case type and difficulty. The study participants had to provide their diagnoses both \emph{with} and \emph{without} the support of a simulated AI system, according to an asynchronous Wizard-of-Oz protocol~\cite{dahlbdck1993wizard}: the support of the AI system included both a proposed diagnosis and a textual explanation to back the former one. The experiment was performed by means of a web-based questionnaire set up through the LimeSurvey platform (version 3.23), to which the readers had been individually invited by personal email.

\begin{figure}[!htb]
  \centering
  \includegraphics[width=0.8\textwidth]{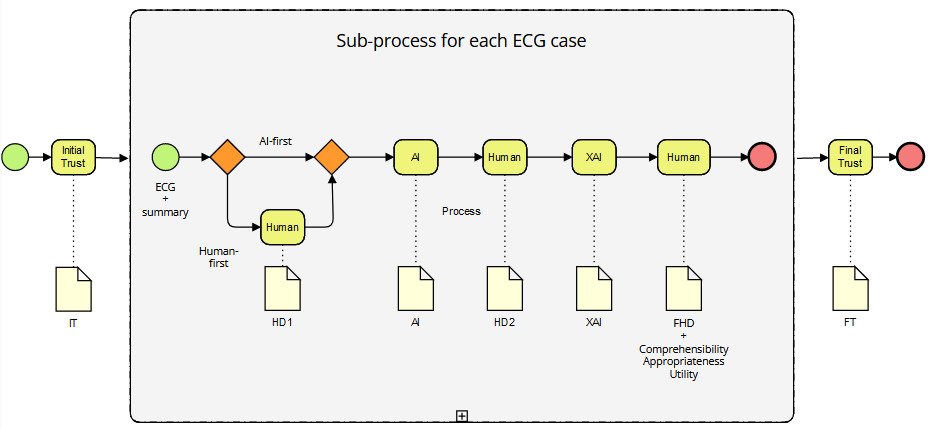}
  \caption{A BPMN representation of the study design. Information collected are represented as data objects, coming from collection tasks, whose name is denoted with the name of the main actor. After the initial collection of the perceived ``trust in AI'', the subprocess is repeated for each ECG case, where HD1, AI, HD2, XAI and FHD items are collected, together with comprehensibility, appropriateness and utility. Finally, a post-test ``trust in AI'' is collected again.}
  \label{fig:protocol}
\end{figure}

The ECG readers were randomly divided in two groups, which were equivalent for expertise and were supposed to interact with the AI system differently (see Fig.~\ref{fig:protocol}); in doing so, we could comparatively evaluate potential differences between a human-first and an AI-first configuration. In both groups, the first question of the questionnaire asked the readers to self-assess their trust in AI-based diagnostic support systems for ECG reading. The same question was also repeated at the end of the questionnaire, to evaluate potential differences in trust caused by the interaction with the AI system.

For each ECG case, the readers in the human-first group were first shown the trace of the ECG together with a brief case description, and then they had to provide an initial diagnosis (in free text format). After that this diagnosis had been recorded, these respondents were then shown the diagnosis proposed by the AI; after having considered this latter advice, the respondents could revise their initial diagnosis; then they were shown the textual explanation (motivating the AI advice) and asked to provide their final diagnosis in light of this additional information. In contrast, the participants enrolled in the AI-first group were shown the AI-proposed diagnosis together with the ECG trace and case description; only afterwards, they were asked to provide their own diagnosis in light of this advice only. Finally, ECG readers were shown the textual explanation, and asked whether they wanted to revise their initial diagnoses or confirm it.

For each textual explanation, we asked the participants to rate its quality in terms of its \emph{comprehensibility}, \emph{appropriateness} and \emph{utility}, in this order (so as to reflect a natural sequence through perception, interpretation and action). 
In particular, while comprehensibility and utility were considered self-explanatory terms, we pointed out in a written comment that appropriateness, to our research aims, would combine the respondents' perception of pertinence and correctness together (that is, that dimension would reflect the extent ``the explanation had something to do with the given advice'' and, \emph{with regard to the latter}, ``it was plausible and correct''\footnote{In other words, we asked the participants to judge the quality of the explanation with respect to the advice and to the case at hand by means of two different constructs (appropriateness and utility, respectively).}).

The accuracy of the simulated AI, that is the proportion of correct diagnostic advices, was 70\%, with respect to the ECG Wave-Maven gold standard. To avoid negative priming, and hence avoid fostering unnecessary distrust in the AI, the first five cases of the questionnaire shown to the participants were all associated with a correct diagnosis and a correct explanation from the XAI support. 
Although the participants had been told that the explanations were automatically generated by the AI system, like the diagnostic advice, these had been prepared by a cardiologist: in particular, 40\% of the explanations were incorrect or not completely pertinent to the cases\footnote{More precisely, for the 5 cases classified as `simple', all explanations were correct; for the 9 cases of medium complexity, 4 explanations were wrong; for the remaining 6 cases denoted as `difficult', 4 explanations were wrong.}.

Based on the collected data, we then considered the following research questions:
\begin{enumerate}[(RQ1)]
  \item Does the readers' expertise have any effect in terms of either basal trust, difference in trust by the readers, or final trust? (RQ1a) Does the interaction protocol have any effect on difference in trust by the readers, or final trust? (RQ1b); 
  \item Is there any difference or correlation between the three investigated psychometric dimensions (i.e. comprehensibility, appropriateness, utility)? A positive answer to this latter question would justify the use of a latent quality construct (defined as the average of the psychometric dimensions) to simplify the treatment of other research questions; 
  \item Do the readers' expertise, their diagnostic ability (i.e., accuracy), as well as the adopted interaction protocol, have any effect in terms of differences in perceived explanations' quality? Similarly, does the perceived explanations' quality correlate with the basal or final trust? In regard to diagnostic ability, we stratified readers based on whether their baseline accuracy (cfr. HD1, see Fig~\ref{fig:protocol}) was either higher or lower than the median; 
  \item Is there any correlation between explanations' perceived quality and the readers' susceptibility to \emph{technology dominance}~\cite{arnold1998theory,sutton2022extension}? Although technological dominance is a multi-factorial concept, for our practical aims, we express it in terms of the rate of decision change due to the exposition to the output of an AI system. Moreover, we distinguished between \emph{positive dominance}, when changes occur from initial wrong decisions (e.g., diagnoses) to eventually correct ones; and \emph{negative dominance}, for the dual case, when the AI support misleads decision makers; 
  \item Finally, does the correctness of the explanation make any difference in terms of either perceived explanations' quality or influence (i.e., dominance)? 
\end{enumerate}

The above mentioned research questions were evaluated by means of a statistical hypothesis testing approach. In particular, correlations were evaluated by means of Spearman $\rho$ (and associated p-values), so as to properly take into account for monotone relationships between ordinal variables and continuous ones. In regard to research questions 1 and 3, on the other hand, paired comparisons were performed by applying Wilcoxon signed-rank test, while un-paired comparisons were performed by applying the Mann-Whitney U test. In both cases, effect sizes were evaluated through the rank biserial correlation (RBC). In all cases, to control the false discovery rate due to multiple hypothesis testing, we adjusted the observed p-values using the Benjamini-Hochberg procedure. Significance was evaluated at the 95\% confidence level.

\section{Results}
After having closed the survey, we collected a total of 1352 responses from the 44 ECG readers involved, of which 21 had been enrolled in the human-first protocol and the remaining 23 in the AI-first protocol.

\begin{figure}[!htb]
  \centering
  \includegraphics[width=0.8\textwidth]{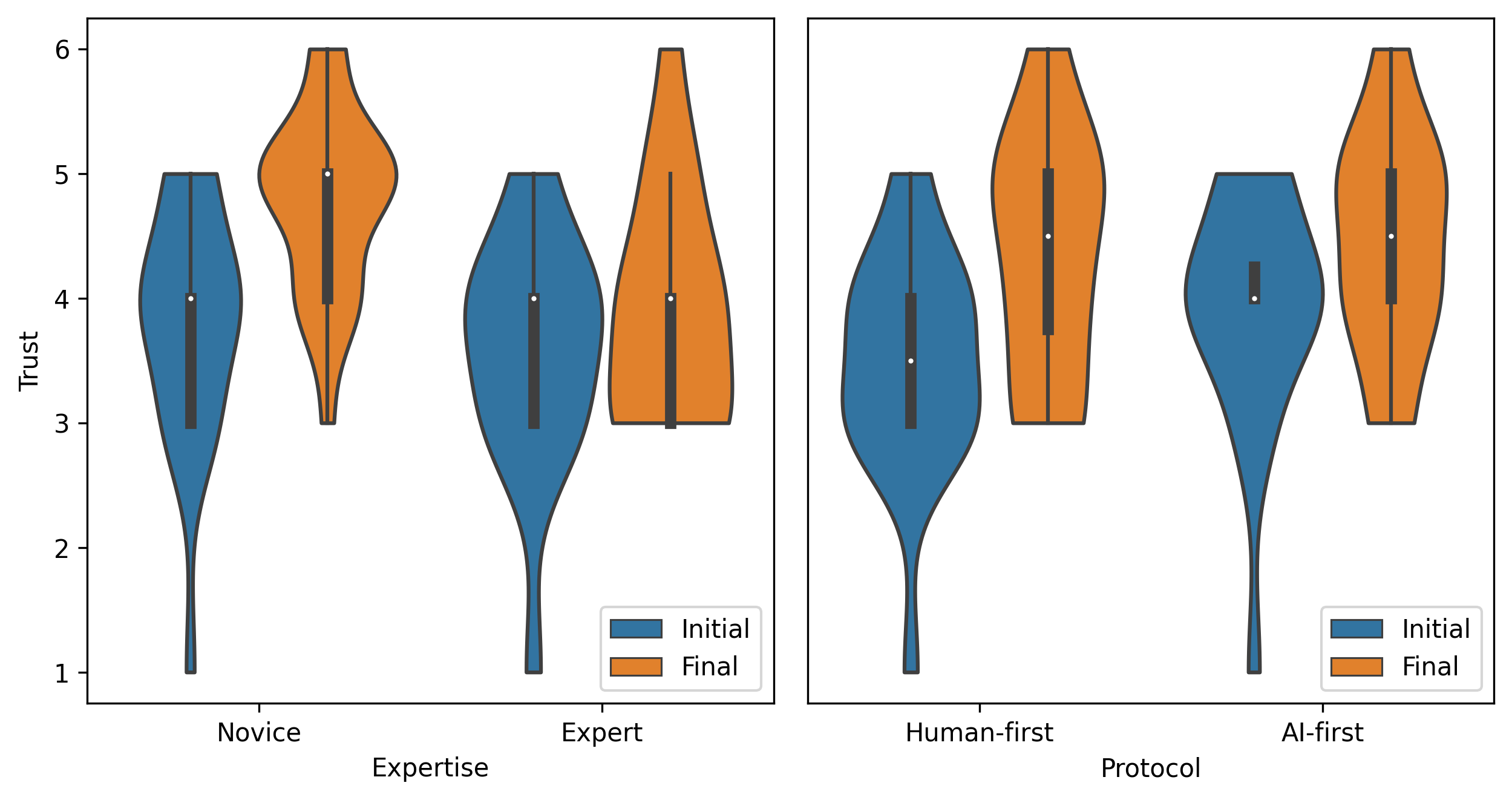}
  \caption{Initial and final trust, stratified by: (left) readers' expertise, (right) interaction protocol.}
  \label{fig:trust}
\end{figure}

The results concerning the differences in self-perceived trust are reported in Figure~\ref{fig:trust}, stratified by expertise (on the left) and interaction protocol (on the right).

The difference between initial and final trust was significant for the novice readers (adjusted p: .004, RBC: 0.92), but not for the expert ones (adjusted p: .407, RBC: 0.35). Furthermore, even though the difference in initial trust between novice and expert readers was not significant (adjusted p: .439, RBC: 0.15), the difference in final trust was instead significant (adjusted p: .009, RBC: 0.54), with the novice readers reporting a higher final trust on average than the expert ones. In regard to the stratification by interaction protocol, the difference between initial and final trust was significant for the human-first cohort (adjusted p: .016, RBC: 0.81) but not so for the AI-first cohort (adjusted p: .078, RBC: 0.60), even though the effect size was large. Differences between the two cohorts were not significant, neither in terms of initial trust (adjusted p: .090, RBC: 0.33) nor in terms of final trust (adjusted p: .787, RBC: 0.05).

\begin{figure}[!htb]
  \centering
  \includegraphics[width=0.8\textwidth]{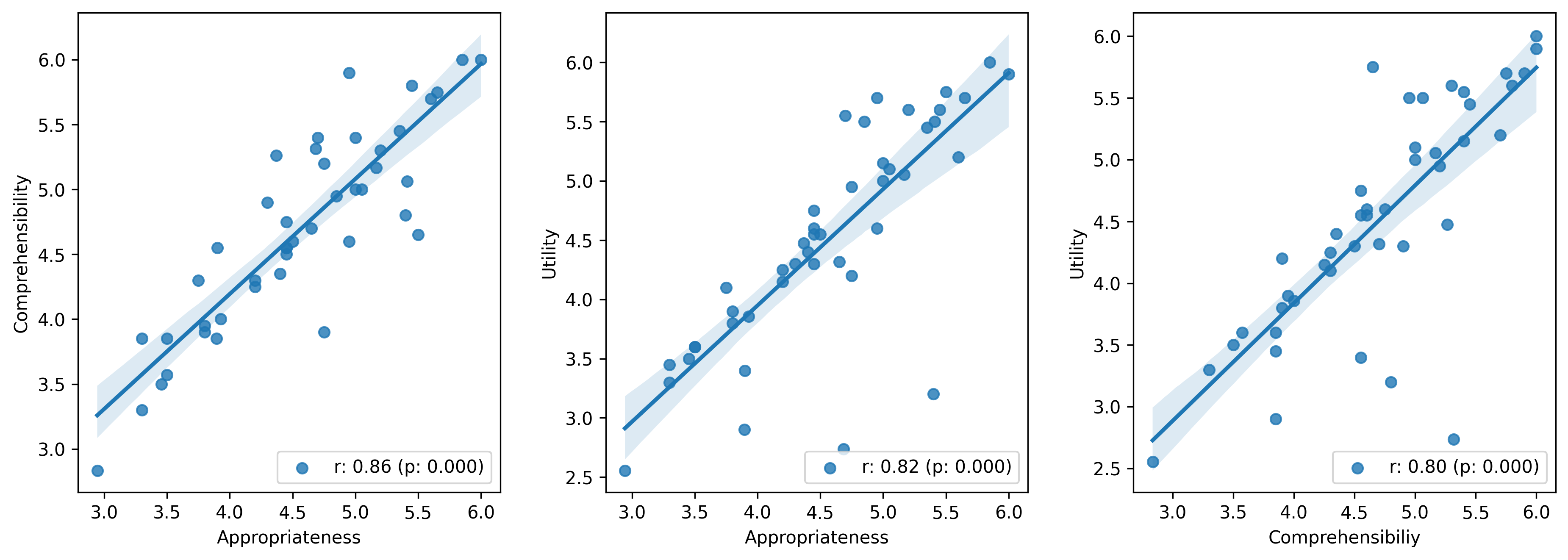}
  \caption{Correlations between appropriateness, comprehensibility and utility.}
  \label{fig:corrs_quali}
\end{figure}

The correlations between the three psychometric dimensions is reported in Figure \ref{fig:corrs_quali}. The three dimensions were all strongly correlated with each other (appropriateness vs comprehensibility, $\rho$: .86; appropriateness vs utility, $\rho$: .82; comprehensibility vs utility, $\rho$: .80), and all of the correlations were significant (adjusted p-values $< .001$). 
Furthermore, the internal consistency of the questionnaire, in regard to the psychometric items (i.e. appropriateness, comprehensibility and utility), was very high (Cronbach $\alpha$: .93). This result suggests that the three psychometric dimensions can be associated with an aggregated quality construct (defined as the average between appropriateness, comprehensibility and utility), which we will consider in what follows.

\begin{figure}[!htb]
  \centering
  \includegraphics[width=0.8\textwidth]{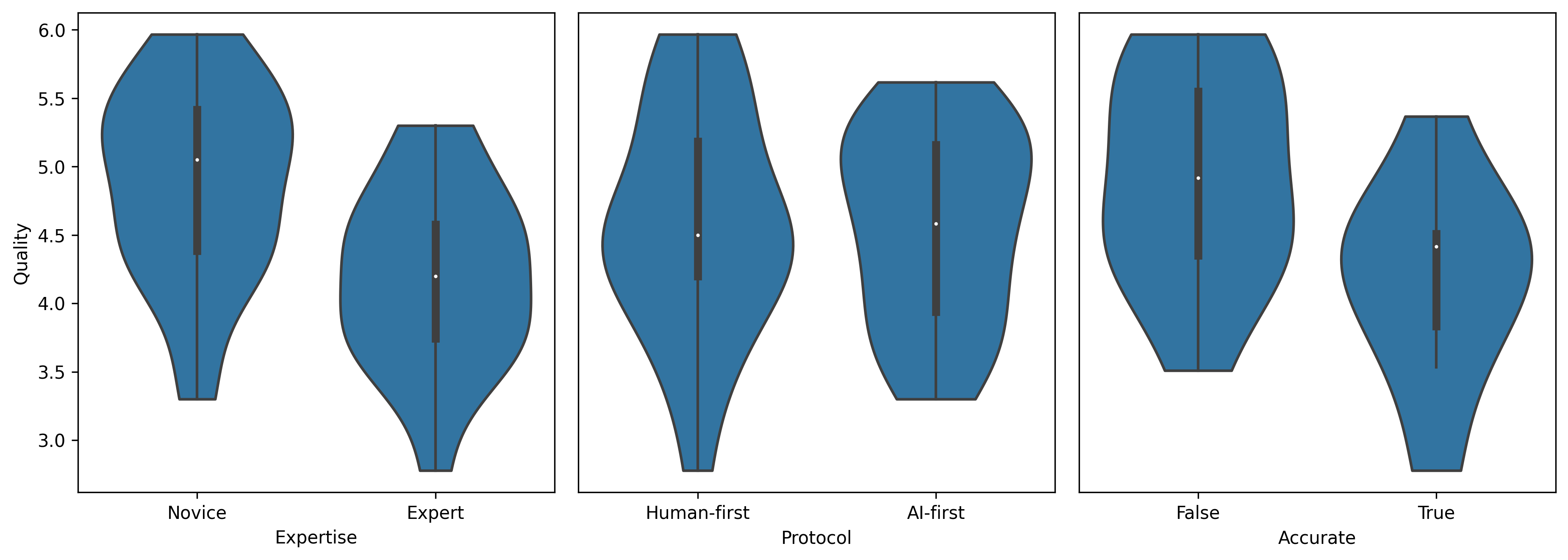}
  \caption{Explanations' quality, stratified by: (left) readers' expertise, (center) interaction protocol, (right) readers' baseline accuracy.}
  \label{fig:quali}
\end{figure}

The results regarding the differences in the perceived quality of the explanations, stratified by either expertise, interaction protocol or readers' baseline accuracy, are reported in Figure~\ref{fig:quali}.

\begin{figure}[!htb]
  \centering
  \includegraphics[width=0.8\textwidth]{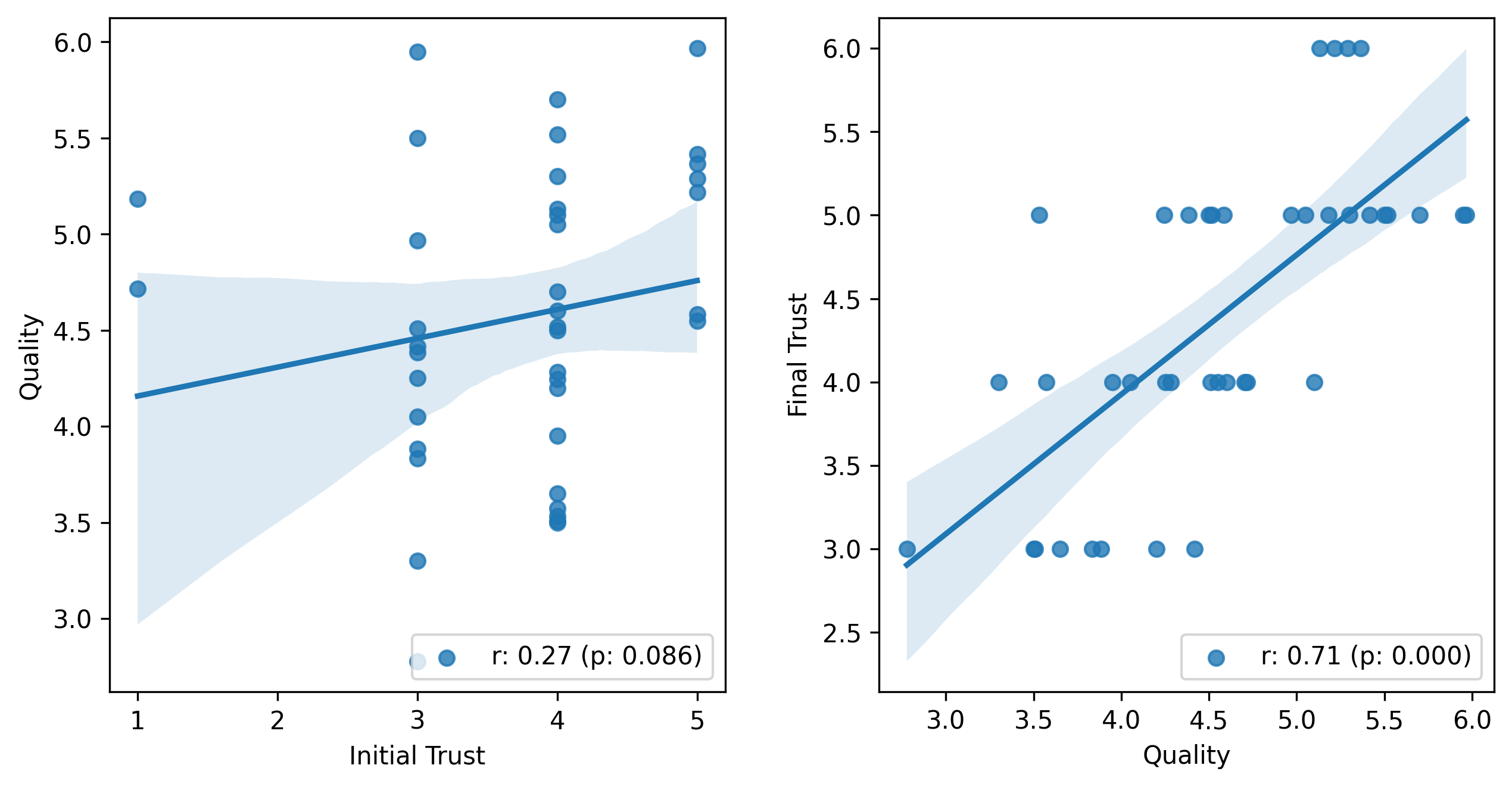}
  \caption{Correlations between perceived explanations' quality and: (left) initial trust, (right) final trust.}
  \label{fig:qualitrust}
\end{figure}

The difference in explanations' quality between human-first and AI-first interaction protocols (adjusted p: .981, RBC: 0.01) was not significant and associated with a negligible effect. Even though the difference in explanations' quality with respect to readers' baseline accuracy was similarly non-significant (adjusted p: .155), the relationship between the two variables was associated with a medium-to-large effect size (0.36). By contrast, the difference in explanations' quality between novice and expert readers (adjusted p: .012, RBC: 0.51) was significant and associated with a large effect size.
The correlations between initial and final trust and explanations' perceived quality is reported in Figure \ref{fig:qualitrust}. Explanations' quality was weakly correlated with the readers' basal trust in AI-based support systems (Spearman $\rho$: .27, adjusted p: .086), but significantly and strongly correlated with final trust (Spearman $\rho$: .71, adjusted p $< .001$).

\begin{figure}[!htb]
  \centering
  \includegraphics[width=0.8\textwidth]{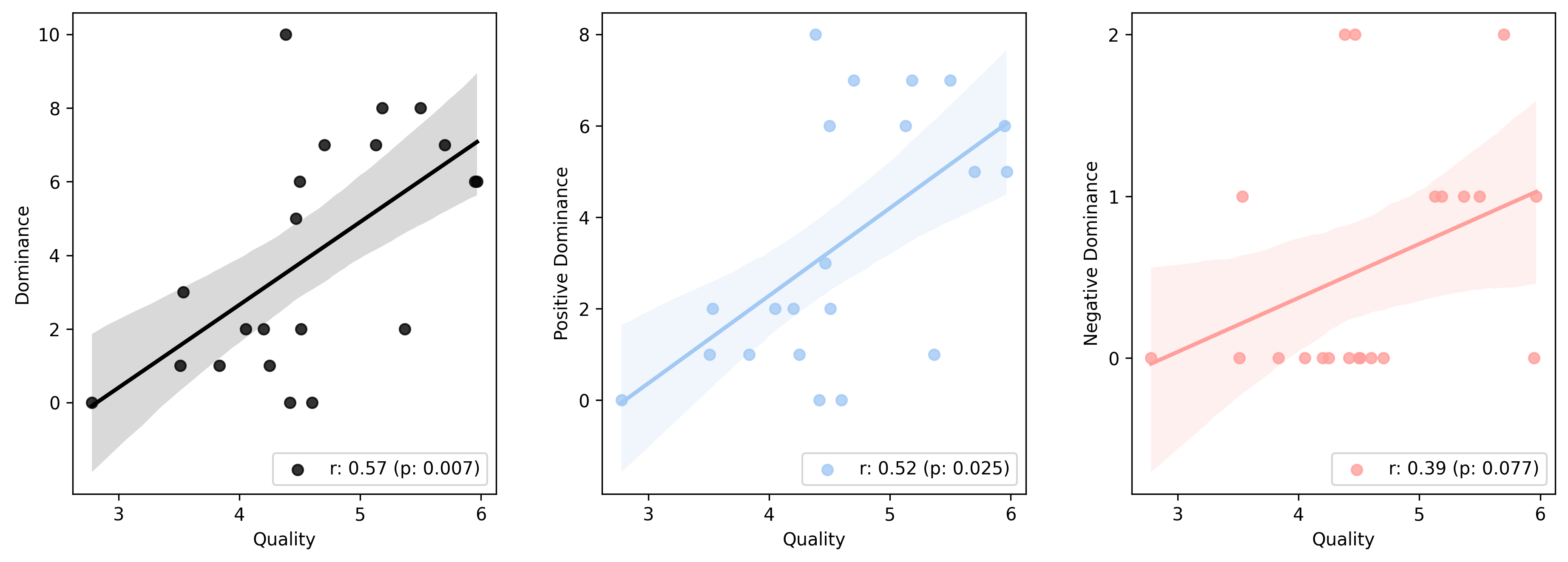}
  \caption{Correlations between dominance (both positive and negative) and explanations' quality.}
  \label{fig:doms}
\end{figure}

The correlations between dominance (distinguishing between positive and negative dominance), and the perceived explanations' quality are reported in Figure~\ref{fig:doms}. Quality was moderately-to-strongly and significantly correlated with dominance (Spearman $\rho$: .57, adjusted: p: .007) and also with the positive component of dominance (Spearman $\rho$: .52, adjusted p: .025); while it was only moderately correlated with negative dominance (Spearman $\rho$: .39, adjusted p: .077); furthermore, this latter correlation was not significant.

\begin{figure}[!htb]
  \centering
  \includegraphics[width=\textwidth]{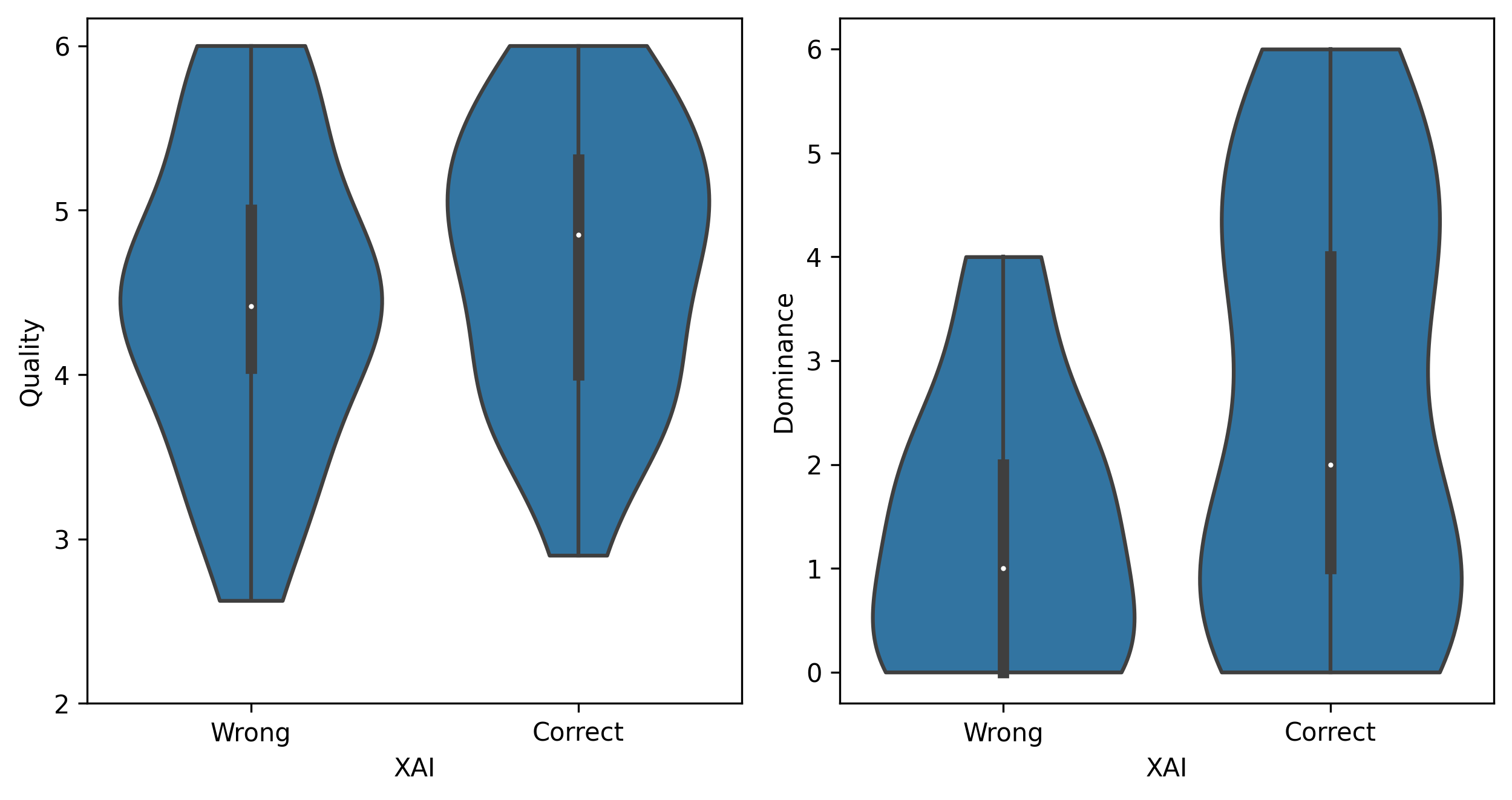}
  \caption{Effect of the explanations' correctness on: (left) perceived explanations' quality, (right) dominance.}
  \label{fig:xai_rels}
\end{figure}

Finally, the relationship between the correctness of the explanations, the perceived quality of these latter, and their dominance is depicted in Figure \ref{fig:xai_rels}. Both the average quality and dominance increased when the explanations were correct, as compared to when these latter ones were wrong: while the observed differences were not significant (quality, adjusted p: .399, RBC: 0.15; dominance, adjusted p: .126, RBC: 0.33), the effects were small-to-medium (for quality) and medium (for dominance).

\section{Discussion}
To discuss the results of the experiment, we follow the order of presentation of the research questions in Section~\ref{sec:methods}. Thus, in regard to RQ1, we point out that we did not observe any significant effect due to the interaction protocol on initial trust. This result does not come unexpected and was in fact desirable: indeed, readers were randomly assigned to one of the two considered protocols. Interestingly yet, the interaction protocol seemingly did not have an effect on final trust either, despite the  fact that interaction protocols had a significant effect in terms of overall accuracy of the hybrid human-AI team \cite{cabitza2022ram}. In this sense, it appears that even though the order of the AI intervention was certainly impactful in regard to the overall diagnostic performance\footnote{This is an effect that is not entirely trivial and calls for further study to understand its causes.}, it might be perceived by the user as a secondary element compared to other trust-inducing or trust-hindering factors. Nonetheless, both interaction protocols had a large effect on trust difference: in particular, human-first protocols led to a significant increase between initial and final trust. A possible explanation for this difference can be found in the accuracy level of the AI system, which in this user study was 70\%, i.e. well over the average accuracy of the readers, and thus likely to lead to a positive interaction for the study participants and hence an increased sense of trust in the decision support. 
More interestingly, even though no significant difference was detected in initial trust between expert and novice readers, the novice readers reported both a significant increase in trust as well as a significantly higher final trust than the expert readers. 
This is in line with previous studies in the field of human-AI interaction \cite{cabitza2021need,cabitza2022color,glick2022impact,paleja2021utility}, which showed how novice readers were more prone to accept the support of an AI-based system, and appreciate its output. An explanation for this widely-reported observation can be traced back to the literature in the \emph{Theory of Technological Dominance} (TTD) \cite{arnold1998theory}, where a previous finding from Noga and Arnold \cite{noga2002tax} identified user expertise as one of the main determinants of dominance and reliance, of which trust is a determinant. While a tenet of TTD holds that decision aids are especially beneficial to professionals thanks to a bias mitigation effect \cite{arnold2004impact}, the study by Jensen et al. \cite{jensen2010technology} displayed a diverging beneficial effect of decision support, with novices benefiting more in comparison to experts who, in turn, often discounted the aid's support. This is in line with our findings, that point to more experienced decision-makers possibly being less favourably impacted by such systems, possibly due to a lower level of familiarity or a higher prejudice against the machine (see also~\cite{cabitza2019biases}).

In regard to RQ2, we briefly note that the three psychometric dimensions that we investigated were indeed strongly correlated between each other. This result, while interesting, is not totally unexpected since, intuitively, an appropriate and comprehensible explanation is likely to be found also useful; conversely, for an explanation to be useful it should be at least also comprehensible. Notably, the observed value of the Cronbach $\alpha$ was higher than Nunnally's reliability threshold for applied studies (i.e. .8, see \cite{nunnally1994psychometric}): thus, the internal consistency of our test was sufficiently high to guarantee its reliability, but not so much as to suggest redundancy and hence undermine its validity \cite{cho2015cronbach}. In particular, we believe that these results justify the aggregation of the three psychometric dimensions into a latent \emph{quality} construct, which was then considered in the statistical analysis.

This perceived quality level was analyzed in RQ3 with respect to user expertise, their accuracy, and the interaction protocol. As expected, the difference in explanations' quality between human-first and AI-first interaction protocols 
was not significant and was associated with only a negligible effect: indeed, as mentioned previously in regard to trust, readers were assigned randomly to the two interaction protocols which, aside from when the AI advice was shown, were essentially equivalent in terms of the given explanations. By contrast, even though the difference in explanation accuracy with respect to readers' baseline accuracy was similarly non-significant,
 the relationship between the two variables was associated with a medium-to-large effect size. Furthermore, also the difference in explanations' quality between novice and expert readers 
 was associated with a large effect size and was also statistically significant. These results highlight how readers' proficiency in the ECG reading task (as measured by either self-reported expertise or, more quantitatively, by basal accuracy) might have a significant effect on the perception of explanatory advice. A possible explanation for this observation, which was already mentioned above in reference to trust, might be related to an increased acquaintance with AI and XAI systems for the less expert readers (who presumably were also less accurate). Furthermore, less expert readers (e.g., the students and residents) might have found the explanations' quality higher due to their perceived usefulness in helping them identify characteristics of interest in an ECG that they were not able to interpret alone: more experienced or more accurate readers, who by definition were more well versed in the interpretation of ECG diagrams, might have missed this novelty element of explanations, which might have led to a lower, on average, perceived quality. Furthermore, explanations' quality was weakly correlated with the readers' basal trust in AI-based support systems, 
but significantly and strongly correlated with final trust. 
Importantly, it appears that the users' initial attitude had little influence on the perceived quality of explanations: this can be interpreted as a consequence of the fact that the explanations were evaluated for their intrinsic value, rather than for a \emph{halo} effect (due to trust)~\cite{brill2019siri}.
As for final trust, by contrast, the observed improvement is likely due to a simple reason: the support was perceived as a worthy addition to the decision-making process. Even though due to how the user experiment was constructed we cannot decouple the contribution of XAI from plain AI, given that the accuracy of the decision support was higher than that of the readers but overall not particularly high (equalling 70\%), we conjecture that the explanation was probably the main factor in the trust increase, for its novelty and appropriability~\cite{yang2022benefiting}
when compared to a simple categorical advice. Since the increase in final trust is highly dependent, and significantly so, on explanation quality, this finding reinforces the idea that explanations do influence trust. We believe that this relevant finding also provides an alternative and complementary explanation of the observed effect of readers' expertise on final trust and trust difference. Indeed, this latter effect could be explained as rising from the fact that less experienced readers rated more favorably the quality of explanations than the more expert readers: in light of the strong relationship between explanations' quality and final trust, this might explain why we observed a larger increase in trust for novices than for experts.

The two final research questions - RQ4 and RQ5 - further delve on the effect of explanations, by adopting the lens of the theory of technology dominance~\cite{arnold1998theory,sutton2022extension} through which we investigated possible correlations between the readers' susceptibility to this phenomenon (which, in this article, was operationalized through the rate of decision change due to the exposition to the output of an AI system) and explanations' perceived quality and actual correctness.

As clearly shown in Figure \ref{fig:doms}, explanation quality influences dominance, and especially positive dominance, strongly and also significantly so. A further confirmation of this effect can also be traced back to the observations in Figure \ref{fig:xai_rels}. Indeed, both quality and dominance were increased when an explanation was correct and pertinent to the cases at hand. We believe this finding to be of particular interest since it confirms that high-quality explanations increase the persuasion potential of the XAI system - and especially so for the good (see, in particular, the rightmost panel in Figure \ref{fig:xai_rels}). Nonetheless, as it can be noticed from the rightmost panel in Figure \ref{fig:doms}, quality can influence the users also for worse, as highlighted by the fact that explanations' quality was moderately associated with negative dominance (hence, opinion changes from a correct to an incorrect diagnosis). A possible explanation for this effect can be traced back to the imperfect ability of the readers' in discriminating a correct explanation from a wrong one in terms of perceived quality (see Figure \ref{fig:xai_rels}). In turn, such an effect could motivate the emergence of biases and cognitive effects associated with automation, and especially so with those that are directly related to the role of XAI, as in the case of the white-box paradox~\cite{cabitza2021need,cabitza2022color}. Despite the relevance of these results, we remark however that further research should address whether this correlation also holds in the case of placebic information (i.e. not semantically sensible, nor structurally consistent) as described by Langer \cite{langer1978mindlessness}, translating this effort from the interpersonal dimension to that of Human-Computer Interaction.




This study is exploratory as its main limitations regard the relatively small sample of cases considered, if not of readers involved. In fact, in regard to participation this study can leverage the perceptions and opinions of tens of cardiology, of different competences and expertise. However, the study regards a serious game where the doctors involved knew no harm could be caused to real patients.
That said, two main areas where further research could extend similar studies regard the stratification by explanation types, and the analysis of the impact of explanations on the readers' confidence. On one hand, explanations should be distinguished according to a reference taxonomy, for instance those recently proposed in~\cite{miller2019explanation,arrieta2020explainable,vilone2021notions,cabitza2022quod}, to see if different types of explanations can have different effects on decision making: we recall that in this study we focused on \emph{textual explanations} of \emph{justificatory} and \emph{causal} kind~\cite{cabitza2022quod}. Moreover, explanations can be wrong in different ways: for instance, an explanation can be wrong because it does not regard (or is badly fitted to) either the case at hand or the machine's advice; or because it expresses a wrong way of reasoning. This macro distinction reflects the typology proposed by Reason about human error~\cite{reason2000human}, in which \emph{lapses} and \emph{mistakes}, respectively, regard errors in perception or attention, and the latter ones regard errors in reasoning and the application of domain knowledge. The explanations provided in the study presented in this paper were of various kinds, depending on the case at hand and the ECG to read.

On the other hand, as mentioned above, we did not collect confidence scores at each human decision step (i.e., HD1, HD2 and FHD, see Figure~\ref{fig:protocol}); for this reason, we cannot address the research question whether explanations would improve confidence in the decision reported or not. However, we addressed this question in another study \cite{RIF}, where preliminary results suggest that (visual) explanations may paradoxically make users (slightly) \emph{less} confident in their final decision.
For this reason, further research should be aimed at investigating also the confidence construct, and its relationship with perceived user experience and satisfaction.

\section{Conclusions}

The current interest in XAI 
is ostensibly and programmatically motivated by the need to make artificial intelligence (AI) systems more transparent, understandable, and thus usable. However, in light of some empirically-grounded findings and the literature on naturalistic decision making~\cite{klein2019naturalistic, gunning2019darpa}, this interest appears to be more instrumental to the rising prevalence and diffusion of \emph{automated decision making} (ADM) systems, especially when their use is anticipated in contexts for which the main legislative frameworks (e.g., the EU GDPR) require these systems to also provide reasons of their output whenever this latter can have a legal effects. This addresses a requirement for justification, rather than explanation, although these two concepts are often conflated (for an argumentation about the difference between explanation and justification, see~\cite{cabitza2022quod}).

In this light, it is important to notice that, metaphorically speaking, providing AI with \emph{explainability}, that is the \emph{capability to properly explain its own output}, is more akin to painting the black box of inscrutable algorithms (such as deep learning or ensemble models) white, rather than making them transparent. What we mean with this metaphoric statement is that XAI explanations do not necessarily \emph{explain} (as by definition or ontological status) but rather describe the main output of systems aimed at supporting (or making) decisions: this is why we described XAI explanations as a \emph{meta output}. As such, explanations can fail to make the output (they relate to) more comprehensible, or its reasons explicit; or, even, they can be wrong. 

This trivial observation is seldom emphasized, and it should motivate researchers to further investigate what happens when these failures occur, not only in terms of AI effectiveness (that is accuracy of the \emph{hybrid decision making}~\cite{reverberi2022experimental}), but also in terms of efficiency (e.g., ``does providing explanations make decision making more time consuming or just more difficult''); and satisfaction (``do the decision makers find the additional information useful, or at least do they feel more confident in their decision after consuming an explanation?'').

Our findings suggest that we should not take the perceived and actual utility of explainable systems for granted: these qualities should be assessed in an ongoing manner, and \emph{in vivo} rather than \emph{in labo}, ensuring that quantitative measures of performance such as error rates, throughput and execution times do not take undue precedence over the evaluation of use experience~\cite{holzinger2021toward}.
Rather, we should ground our design choice to make AI systems explainable (that is, capable of supplying explanations) on empirical evidence on the basis of the \emph{fit} (or cognitive congruence) between the user and the artifact (i.e., trust and expectations), user and task (i.e., skill-difficulty match, expertise), and the artifact and the task to support, as proposed in the theory of dominance~\cite{arnold1998theory,sutton2022extension}. 

By evaluating the quality and usefulness of explanations in relation to user perception and performance, rather than in isolation, our study brought to light some paradoxical effects related to the introduction of explanations into diagnostic AI systems; therefore our study aims at contributing to the discussion around the necessity of a \emph{relational} approach to AI design and evaluation. Following Virginia Dignum \cite{dignum2022relational}, we also call for greater attention to the dynamics of decision-making settings, as well as to how humans and machines come to interact, and even ``collaborate'' in so-called Hybrid Human-Artificial Intelligence ensembles~\cite{reverberi2022experimental, dellermann2019future}.
This leads to our belief that, even more so than from Engineering and Computer Science, the greatest advances for AI are likely to emerge from a multidisciplinary effort gathering relevant contributions from the scholarly fields of Cognitive Ergonomics~\cite{andrews2022role,neerincx2018using}, Social psychology~\cite{cooke2021effective}, Human-Computer Interaction~\cite{shneiderman2020human}, Computer-Supported Cooperative Work~\cite{liu2021ai,wang2020human}, and Human Factors~\cite{asan2021research,parasuraman2000model}.

\end{document}